\documentclass[10pt,twocolumn,letterpaper]{article}

\usepackage{iccv}
\usepackage{times}
\usepackage{epsfig}
\usepackage{graphicx}
\usepackage{amsmath}
\usepackage{amssymb}

\usepackage{authblk}
\usepackage{xcolor}
\usepackage{blindtext}
\usepackage{bbm}
\usepackage{enumitem}
\usepackage{epsfig}
\usepackage{graphicx}
\usepackage{amssymb}
\usepackage[ruled]{algorithm2e}
\usepackage{algorithmic}
\usepackage{multirow}
\usepackage{color}
\usepackage{color, colortbl}
\usepackage{array}
\usepackage[english]{babel} % English language/hyphenation
\usepackage{amsmath,amsfonts,amsthm} % Math packages
\usepackage{mathtools}
\usepackage{pifont}
\newcommand{\xmark}{\ding{55}}
\usepackage[pagebackref=true,breaklinks=true,letterpaper=true,colorlinks,bookmarks=false,breaklinks=true]{hyperref}

\iccvfinalcopy % *** Uncomment this line for the final submission

\def\iccvPaperID{****} % *** Enter the ICCV Paper ID here

\ificcvfinal\fi
\begin{document}

%%%%%%%%% TITLE
\title{Embarrassingly Simple Binary Representation Learning}

\author[1]{Yuming Shen
}
\author[1]{Jie Qin
}
\author[1]{Jiaxin Chen
}
\author[1]{Li Liu
}
\author[1]{Fan Zhu
}
\vspace{-3ex}
\affil[1]{Inception Institute of Artificial Intelligence (IIAI), Abu Dhabi, UAE}
\affil[ ]{\small \texttt{\{ymcidence, qinjiebuaa, chenjiaxinx, liuli1213,  fanzhu1987\}@gmail.com}}

\maketitle
\thispagestyle{empty}

%%%%%%%%% ABSTRACT
\begin{abstract}
   Recent binary representation learning models usually require sophisticated binary optimization, similarity measure or even generative models as auxiliaries. However, one may wonder whether these non-trivial components are needed to formulate practical and effective hashing models. 
   
   In this paper, we answer the above question by proposing an embarrassingly simple approach to binary representation learning. %Optimized 
   With a simple classification %negative log-likelihood 
   objective, our model only incorporates two additional fully-connected layers onto the top of an arbitrary backbone network,%for binary latents and semantic labels respectively,
   whilst complying with the binary constraints during training. The proposed model lower-bounds the Information Bottleneck (IB) between data samples and their semantics, and can be related to many recent `learning to hash' paradigms. We show that, when properly designed, even such a simple network can generate effective binary codes, by fully exploring data semantics without any held-out alternating updating steps or auxiliary models. Experiments are conducted on conventional large-scale benchmarks, i.e., CIFAR-10, NUS-WIDE, and ImageNet, where the proposed simple model outperforms the state-of-the-art methods. Our codes are available at \url{https://github.com/ymcidence/JMLH}.
\end{abstract}

\section{Introduction}
Approximate nearest neighbour search with binary representations has been regarded as an effective and efficient solution to large-scale multimedia data retrieval. Conventionally termed as \textit{learning to hash}, this family of techniques aims at \textbf{(a)} shrinking the embedding size of data and \textbf{(b)} producing binary features to speedup the computation of distance-based pair-wise data relevance. Similar to many other machine learning tasks, learning to hash can be either unsupervised or supervised. The former requires less labeling efforts for training, while the later obtains better performance in retrieval. We focus on supervised hashing to fully leverage the semantic information of data.

Recent research in this field largely boosts the performance of the produced hash codes by introducing deep learning techniques. Deep hashing models typically employ an indifferentiable \texttt{sign} activation to the top of the encoding network. Various methods have been proposed to empower the encoder with the ability to properly locate data in the Hamming space. 

A typical approach is to employ a held-out code learner as the network training complementary \cite{bdnn,sdh,zsh}. The code learner performs discrete optimization and alternately updates the semantic-based target codes to govern the behavior of the encoding network. This approach generally requires longer training time since the held-out discrete optimization step cannot be effectively paralleled, and consumes additional memory to cache the target codes during each round of update. Alternatively, some propose to decouple unrelated data representations by introducing similarity-based penalties to the encoders \cite{Cao_2017_ICCV,gcnh,dhn,zhuang2016fast}. To train an encoder with these regularizers, one may resort to continuous relaxation on the codes, which arguably degrades the training quality. One recent fashion in deep hashing is to employ generative adversarial models \cite{shashgan,hashgan,bgan,bingan}. By distinguishing synthesized data from real ones, the encoder implicitly acknowledges the respective data distribution.

However, the above precisely-proposed approaches raise another question: \emph{%Does it really take so much effort to build an effective hashing model?
	How to build an effective supervised hashing model with \textbf{minimum} auxiliary components?}

We attempt to find the answer by carefully considering the following main challenges of learning to hash:
\begin{itemize}
	\item \textbf{Keeping the discrete nature of binary codes.} The binary constraints usually lead to an NP-hard optimization problem in parameterized models, and cannot be directly solved by gradient-based methods. This is usually addressed by conventional methods using held-out discrete optimization or relaxation techniques.
	\item \textbf{Enriching the information carried by the codes.} It is always essential to make the encoder aware of the semantic information (\eg, lables or tags) of data.
\end{itemize}

As a result, in this paper, we propose a simple but powerful deep hashing network. In our model, the above problems are tackled by relating data and their semantics with a binary representation bottleneck, which is thereafter used as the final hash codes. A single recognition penalty is applied for training. With a reasonable regularization term, the final learning objective forms a variational lower bound of the Information Bottleneck (IB)~\cite{vib,ib} between observed data and their semantics. Importantly, one can impose stochasticity on the binary bottleneck to keep the binary constraints and apply gradient estimation methods during training. Therefore, the whole framework can be optimized end-to-end with Stochastic Gradient Descent (SGD). To this end, we find our design leads to an embarrassingly simple solution,% namely \textit{recognition} + \textit{SGD}, 
which basically shapes a single classification neural network . 

Regardless of the regularization, the proposed model just maximizes the label likelihood of data. Thus, we name our model Just-Maximizing-Likelihood Hashing (JMLH). The contributions of this paper are summarized as follows:
\begin{itemize}
	\item We propose a simple and novel deep hashing model, \ie, JMLH, and theoretically base it on the Variational Information Bottleneck (VIB)~\cite{vib} method. To the best of our knowledge, JMLH is the first attempt in deep hashing to employ the IB methods.
	\item We show that, when properly designed and trained, a classification neural network with a discrete bottleneck already produces effective binary representations. Therefore, the proposed model requires no auxiliary components and can be optimized directly.
	\item Relations between JMLH and many existing hashing models are discussed in detail.%, in terms of semantics preserving, optimization and the consequences of discrete-continuous relaxation.
	\item JMLH successfully outperforms state-of-the-art hashing techniques on several benchmark datasets, \ie, CIFAR-10~\cite{cifar}, NUS-WIDE~\cite{nus} and ImageNet~\cite{imagenet}.
\end{itemize}

In the rest of this paper, we first describe our model in detail %and the theoretical link with VIB~\cite{vib} 
in Section~\ref{sec_2}. Subsequently, the relationships between JMLH and existing works are elaborated in Section~\ref{sec_3}. Section~\ref{sec_4} presents the implementation details and experimental findings, with a brief conclusion given in Section~\ref{sec_5}.
%-------------------------------------------------------------------------
\section{Model}~\label{sec_2}
The goal of learning to hash is to find an optimal encoding function $f:X\rightarrow\{0,1\}^m$ to represent data. Here $X$ is the variable space of data observation and $m$ refers to the length of the hash code space $B$. In the context of supervised hashing, training is usually supported by the data labels $Y$. We intendedly use capitalized notations, \ie, $X$, $Y$ and $B$, for the (random) variable spaces, and denote each respective variable instances with lower-cased letters, \ie, $\mathbf{x}$, $y$ and $\mathbf{b}$.

\subsection{JMLH at a Glance}
\begin{figure}[t]
	\begin{center}
		\includegraphics[width=0.6\linewidth]{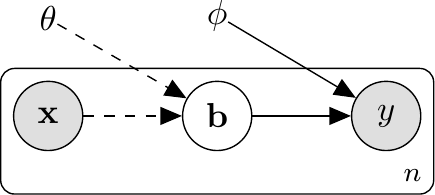}
	\end{center}
	\caption{The directed graphical model of JMLH. We treat the hash code $\mathbf{b}$ as the latent bottleneck between data $\mathbf{x}$ and their labels $y$. The dotted lines define the stochastic encoding procedure of $q(B|X)$, and the solid lines denote the approximated likelihood $q(Y|B)$. $n$ is the total number of observed data points. Note that the respective parameters $\theta$ and $\phi$ are jointly learned, forming an extremely simple training model.}
	\label{fig_1}
\end{figure}

JMLH involves a stochastic encoder $q(B|X)$ and a classifier $q(Y|B)$. An additional deterministic distribution $p(B)$ is used as the prior of $B$.\footnote{Here we use $q\left(\cdot\right)$ to denote an approximated posterior when one cannot directly model the corresponding true distribution, \eg, $q(B|X)$. On the other hand, $p(\cdot)$ is used when the distribution can be deterministically defined or computed, \eg, the pre-defined prior $p(B)$.} This model is illustrated in Figure~\ref{fig_1} as a directed graphical model. 
Particularly, each datum $\mathbf{x}\in X$ is firstly associated with a latent binary code $\mathbf{b}\in B$ according to $q(B|X)$, and then the respective label $y\in Y$ can be predicted by feeding $q(Y|B)$ with $\mathbf{b}$. Therefore, $B$ can be regarded as the bottleneck between $X$ and $Y$. Successively applying $q(B|X)$ and $q(Y|B)$ according to the above procedure specifies \textbf{a single-task neural network} with a binary layer in between, which makes JMLH extremely simple.

We firstly describe the above-mentioned probabilistic models and then discuss how they are combined as a whole for efficient end-to-end training.
\subsubsection{Parameterizing the Probability Models}

Given a training pair of $\left(\mathbf{x},y\right)$, the corresponding probabilities models of $q(\mathbf{b}|\mathbf{x})$ and $q(y|\mathbf{b})$ in JMLH are defined as
\begin{equation}\label{eq_1}
\begin{split}
q(\mathbf{b}|\mathbf{x}) &= \mathcal{P}(\mathbf{b}|\kappa(\mathbf{x};\theta)),\\
q(y|\mathbf{b})&= Cat(y|\pi(\mathbf{b};\phi))\text{~or~}\mathcal{P}(y|\pi(\mathbf{b};\phi)),\\
p(\mathbf{b}) &= \mathcal{B}(\mathbf{b}|m,0.5).
\end{split}
\end{equation}
Here $\mathcal{P}(\mathbf{b}|\kappa(\mathbf{x};\theta))$ indicates the Poisson binomial distribution, parameterized by a neural network $\kappa(\mathbf{x};\theta)$ as follows:
\begin{equation}
\mathcal{P}(\mathbf{b}|\kappa(\mathbf{x};\theta))=\prod_{i=1}^m\kappa_i^{\mathbf{b}_i}(1-\kappa_i)^{1-\mathbf{b}_i}.
\end{equation}
On the other hand, $p(y|\mathbf{b})$ can be either categorical for single-label classification, \ie, $ Cat(y|\pi(\mathbf{b};\phi))$, or Poisson binomial for multi-label classification, \ie, $\mathcal{P}(y|\pi(\mathbf{b};\phi))$, implemented by another network $\pi(\mathbf{b};\phi)$. We additionally introduce $p(\mathbf{b})$ of a binomial distribution $\mathcal{B}(\mathbf{b}|m,0.5)$ as the code prior for regularization purpose.

Note that we choose discrete probability models for $B$ to avoid the use of continuous relaxation. That is to say, the input to the classifier $\pi(\cdot)$ is already binarized. Continuous relaxation, \eg, activating the neurons with a \texttt{sigmoid} non-linearity, is not considered here as it skews the observation of the classifier, propagating biased semantic information measurement back to the encoder.

\subsubsection{Shaping a Single Network}
Sequentially stacking $\kappa(\mathbf{x};\theta)$ and $\pi(\mathbf{b};\phi)$ empirically forms a classification neural network with a binary bottleneck $B$, of which the briefed structure is illustrated in Table~\ref{tab_net}. It can be seen that JMLH only introduces two additional layers on the top of an arbitrary network backbone, which makes it easy to be adopted to different pre-trained models and is convenient for implementation.

Then we define the learning objective with $n$ given training pairs $\{(\mathbf{x},y)\}^n$ of this single network as

\begin{equation}\label{eq_3}
\begin{split}
\mathcal{L} 
= \frac{1}{n}\sum_{(\mathbf{x},y)}\underbrace{\mathbb{E}_{q(\mathbf{b}|\mathbf{x})}[-\log q(y|\mathbf{b})]}_{\text{classification objective}}
+ \underbrace{\lambda \operatorname{KL}\left(q(\mathbf{b}|\mathbf{x})||p(\mathbf{b})\right)}_{\text{regularization}},
\end{split}	
\end{equation}
where $\lambda$ is a hyper-parameter. All the probability models are defined in Eq.~(\ref{eq_1}). We first elaborate each component of it in this subsection and later show that this learning objective is supported by VIB~\cite{vib} in Section~\ref{sec_221}.

\begin{table}[t]
	\caption{Network settings of JMLH. All layers are sequentially applied.}
	\label{tab_net}
	\small
	\resizebox{0.99\linewidth}{!}{
		\begin{tabular}{c c c}
			\rowcolor{gray!20}\hline
			\textbf{Notation}&\textbf{Specification}&\textbf{Variable}\\\hline\hline
			\multirow{2}{*}{Input}& Arbitrary data, &\multirow{2}{*}{$X$}\\
			&$256\times 256$ images in our experiments&\\\hline\hline
			\multirow{5}{*}{$\kappa(\mathbf{x};\theta)$}& Arbitrary network backbone,&\\
			&Alexnet~\cite{alexnet} before \texttt{fc\_7}&\\
			&in our experiments&\\\cline{2-3}
			& Fully-connected, size of $m$& \multirow{2}{*}{$B$}\\
			& Binary stochastic activation&\\\hline\hline
			\multirow{3}{*}{$\pi(\mathbf{b};\phi)$}&Fully-connected, size of label length&\multirow{3}{*}{$Y$}\\
			&\texttt{softmax} (single-label datasets)&\\
			&\texttt{sigmoid} (multi-label datasets)&\\\hline
		\end{tabular}
	}
	\vspace{0ex}
\end{table}

The first Right-Hand-Side (RHS) term of Eq. (\ref{eq_3}), \ie $-\log q(y|\mathbf{b})$, is actually a negative log-likelihood classification penalty since $q(y|\mathbf{b})$ is categorical. This loss conveys semantic label information of data to their codes during training.

The second RHS term of Eq. (\ref{eq_3}) acts as a regularizer. By minimizing the Kullback-Leibler (KL) divergence between the posterior $q(\mathbf{b}|\mathbf{x})$ and the prior $p(\mathbf{b})$, the entropy carried by $B$ is reserved. As the prior and the posterior are basically binomial,the KL divergence can be deterministically computed by two entropy terms $\mathcal{H}(\cdot)$:
\begin{equation}
\operatorname{KL}\left(q(\mathbf{b}|\mathbf{x})||p(\mathbf{b})\right) = 
\mathcal{H}\big(q(\mathbf{b}|\mathbf{x}),p(\mathbf{b})\big) - \mathcal{H}\big(p(\mathbf{b}),p(\mathbf{b})\big).
\end{equation}

The whole network of JMLH is trained only using Eq.~(\ref{eq_3}). This makes the optimization extremely simple, requiring no auxiliary module or additional complex loss function. The only problem comes from the gradient computation of the intractable expected negative log-likelihood \wrt $\theta$, which is discussed in Section~\ref{sec_213}.
\subsubsection{On the Tractability of JMLH}\label{sec_213}
\begin{algorithm}[t]
	\small
	\caption{The Training Procedure of JMLH}
	\label{alg}
	
	\textbf{Input:}\hspace{0mm} Data observations $X$, the corresponding labels $Y$ and the maxinum number of iterations $T$.\\
	\textbf{Output:}\hspace{0mm} Network parameters $\theta$.\\
	%	\BlankLine
	%Randomly initialize $\mathbf{H}\in\{-1,1\}^{M\times N}$\\
	\Repeat{convergence or reaching the maximum iteration $T$}{
		Randomly pick a batch of $\{(\mathbf{x},y)\}$ from training data\\
		Sample $\epsilon\sim\mathcal{U}\left(0,1\right)^m$ for each datum\\
		$\mathcal{L}\leftarrow$ Eq.~(\ref{eq_3})\\
		$(\theta, \phi)\leftarrow\Big(\theta-\Gamma\left(\nabla_\theta\mathcal{L}\right), \phi-\Gamma\left(\nabla_\phi\mathcal{L}\right)\Big)$ according to Eq..~(\ref{eq_6})\\	
	}
\end{algorithm}

Computing the gradients of the negative log-likelihood expectation term $\nabla_\theta\mathbb{E}_{q(\mathbf{b}|\mathbf{x})}\left[-\log q(y|\mathbf{b})\right]$ of Eq.~(\ref{eq_3}) is intractable. One needs to traverse the latent space of $B$ for each sample $\mathbf{x}$ to accurately obtain the loss and corresponding gradients. Inspired by \cite{sgh}, we use the following reparametrization of $B$:\footnote{Although the reparametrization trick~\cite{vae} is initially designed for continuous variables, we keep using this terminology here, because the trick proposed in \cite{sgh} leads to a similar gradient estimator to the one of \cite{vae}.}

\begin{equation}\label{eq_5}
\widetilde{\mathbf{b}}_{i}=
\begin{cases}
1&{\kappa_i(\mathbf{x};\theta)\geqslant\epsilon_i,}\\
0&{\kappa_i(\mathbf{x};\theta)<\epsilon_i,}
\end{cases} \quad\text{for~} i=1~...~m,
\end{equation}
where each $\epsilon_i\sim\mathcal{U}\left(0,1\right)$ is a small random signal. Eq.~(\ref{eq_5}) is conventionally termed as the stochastic binary neural activation. With this reparametrization, the gradient of $\mathcal{L}$ \wrt the encoder parameters $\theta$ can be estimated by the distributional derivative estimator~\cite{sgh}:
\begin{equation}\label{eq_6}
\begin{split}
\nabla_\theta\mathcal{L}=\frac{1}{n}\sum_{(\mathbf{x},y)}\Big(\mathbb{E}_\epsilon[-&\nabla_\theta\log q(y|\widetilde{\mathbf{b}})]\\
&+\lambda\nabla_\theta \operatorname{KL}\left(q(\mathbf{b}|\mathbf{x})||p(\mathbf{b})\right)\Big)
\end{split}
\end{equation}

\begin{figure}[t]
	\begin{center}
		\includegraphics[width=0.75\linewidth]{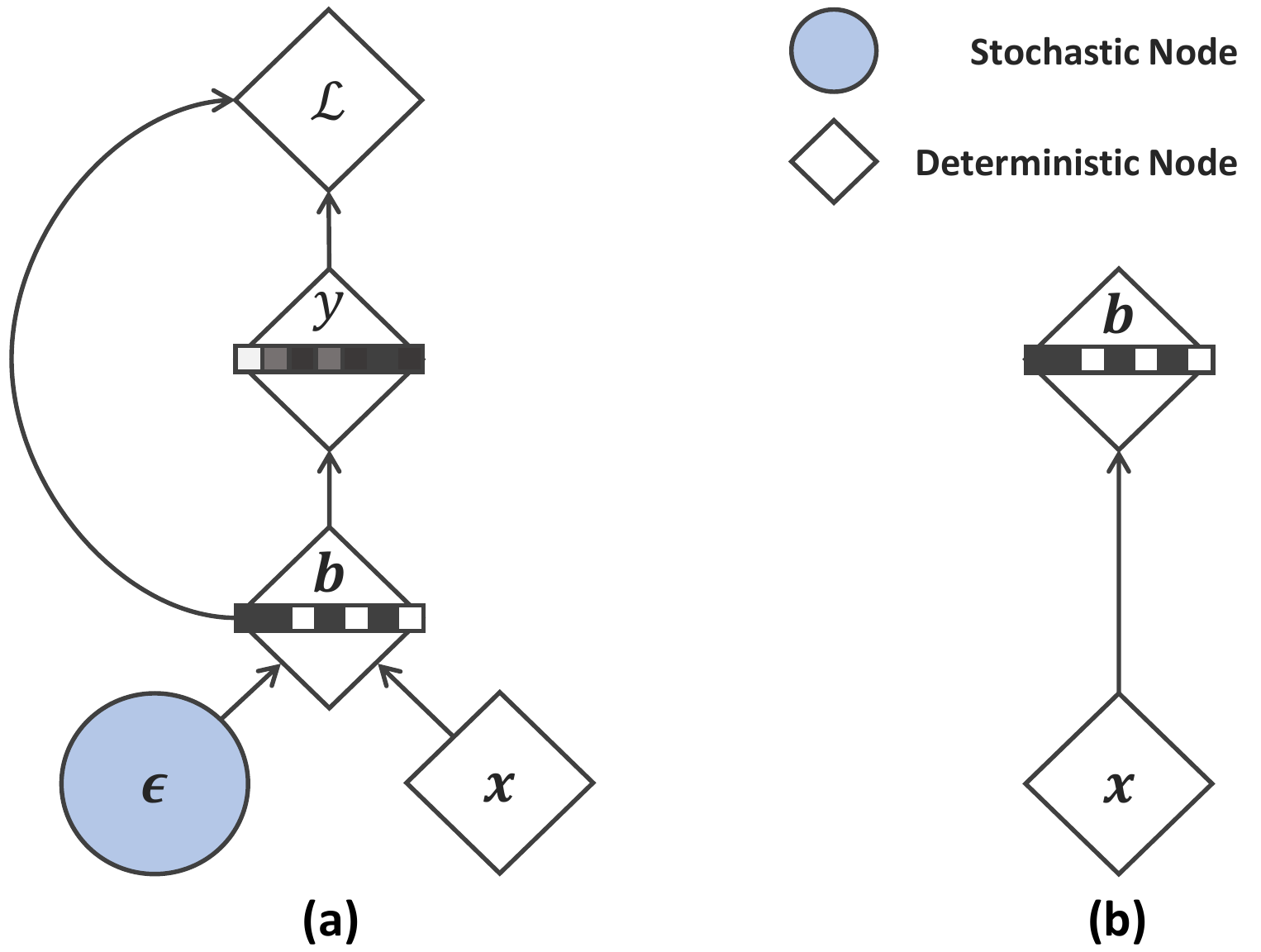}
	\end{center}
	\caption{An analogy of the JMLH computation graphs for \textbf{(a)} training and \textbf{(b)} test.}
	\label{fig_com}
\end{figure}

With this estimator, the network of JMLH can be trained with SGD end-to-end. Note that $\nabla_\phi\mathcal{L}$ can be deterministically obtained and does not require approximation since $\pi(\mathbf{b};\phi)$ does not involve stochasticity. 

The whole training process is illustrated in Algorithm~\ref{alg}, and the respective variable feed path is illustrated in Figure~\ref{fig_com} (a). Here we use $\Gamma(\cdot)$ to denote the gradient scaler, which is the Adam optimizer~\cite{adam} in this work. It can be seen that, during training, JMLH performs identically to a normal neural classifier. The only additional step is just to sample the random signals $\epsilon$.
\subsubsection{Out-of-Sample Extension}
Given a query datum $\mathbf{x}^{(q)}$, the corresponding hash code is produced by the encoder, \ie,
\begin{equation}
\mathbf{b}^{(q)}=\big(\operatorname{sign}(\kappa(\mathbf{x}^{(q)};\theta)-0.5)+1\big)/2,
\end{equation}
which is shown in Figure~\ref{fig_com} (b).
\subsection{Theoretical Analysis}

\subsubsection{Exploring the Information Bottleneck}\label{sec_221}
In this subsection, we show that JMLH defines a special discrete extension of VIB~\cite{vib} to learn information-rich codes. By empirically assigning the joint probability of $X$ and $Y$ with the Dirac delta function  $p(\mathbf{x},y)=\frac{1}{n}\sum_i\delta(\mathbf{x}-\mathbf{x}_i)\delta(y-y_i)=p(y|\mathbf{x})p(\mathbf{x})$, \ie, data samples are independent, the negative learning objective of JMLH can be rewritten as
\begin{equation}
\begin{split}
-\mathcal{L}=\frac{1}{n}\sum_{(\mathbf{x},y)}\sum_\mathbf{b}\Big(p(\mathbf{x})p(y|\mathbf{x})q(\mathbf{b}|\mathbf{x})\log q(y|\mathbf{b})\\
-\lambda p(\mathbf{x})q(\mathbf{b}|\mathbf{x})\log\frac{q(\mathbf{b}|\mathbf{x})}{p(\mathbf{b})}\Big),
\end{split}
\end{equation}
where the first RHS term is the variational lower bound of the mutual information $I(B,Y)$ with the second RHS term the lower bound of the negative mutual information $-\lambda I(B,X)$ according to \cite{vib}. Consequently, $-\mathcal{L}$ literally lower-bounds the IB~\cite{ib} objective $\mathcal{R}_{IB}(X,Y,B)$:
\begin{equation}\label{eq_9}
\mathcal{R}_{IB}(X,Y,B)=I(B,Y)-\lambda I(B,X)\geq -\mathcal{L}.
\end{equation}
We refer to the related articles~\cite{vib,ib} for more detailed definitions.

Intuitively, our learning objective allows $B$ to maximally represent the semantic meaning of the label space $Y$ by ascending $I(B,Y)$. Note that, though $-\lambda I(B,X)$ acts as a penalty in Eq.~(\ref{eq_9}), we are not expecting zero mutual information between $X$ and $B$, otherwise the produced codes would be data-independent. The purpose of introducing $-\lambda I(B,X)$ is to filter redundant information not related to the semantic meanings of data during encoding, and simultaneously preserve the essential part to support $I(B,Y)$. In this way, the learned codes can be compressed and discriminative.

\subsubsection{Nearest Neighbour Search with Recognition}
In the context of large-scale data retrieval, relevant data pairs are usually and conveniently defined by sharing the labels/tags, which is generally reasonable. It is trivial and inefficient to traverse all data points in a dataset and explicitly assign pair-wise similarity marks to each of them, while the labels/tags can be regarded as the similarity `anchors' to ease this process.

JMLH favors this setting as it is literally a special classifier during training. The bottleneck latents $B$ are directly linked to the data labels. When the model is well-trained, the codes of relevant data are naturally located with short Hamming distances. This idea has also been proved in many label-based hashing approaches~\cite{subic,sdh}.

\section{Related Work}~\label{sec_3}
Our work is related to various hashing techniques, of which the most popular and related ones are selectively discussed according to our motivation and design.
\subsection{Solving the Discrete Constraints}
\noindent \textbf{Traditional solutions.} We firstly look at the problem of discrete optimization. A typical example is SDH~\cite{sdh}, which also sequentially behaves encoding and classification. However, as SDH~\cite{sdh} resorts to Discrete Cyclic Coordinate descent (DCC) for alternating code updating, a held-out optimization step is involved. Practically, this is hard for parallelization and batch-wise optimization. Additionally, training errors of the classification step cannot be efficiently propagated back to the encoder. A similar paradigm can be found in~\cite{cnnh}, while its objective is based on pair-wise data similarity. In both single-modal hashing~\cite{zsh,bdnn} and cross-modal hashing~\cite{dsh,mimi}, alternating code updating is widely adopted. %Notably, a great number of deep hashing techniques introduces binary-continuous relaxation on the encoder output layer with a \texttt{sigmoid} or \texttt{tanh} non-linearity. 
For those methods that have held-out code-learners, the network is regularized by the produced target code. The disadvantage of this disarticulated process is the low training quality. On the other hand, regularizing the network by quantization is also widely considered~\cite{dch,dh,subic,dvbj}. However, these approaches ignore a severe problem of the different presence of codes. The network observes continuous codes during training, which may represent different meanings from their discrete counterparts for test. This problem is explicitly solved in JMLH as our code bottleneck is exactly binary.

\noindent \textbf{Gradient estimation solutions.} Some existing hashing models solve the discrete constraints for SGD by gradient estimation techniques so that the hashing model can be conveniently trained. In SGH \cite{sgh}, a distributional derivative estimator is proposed based on the Taylor expansion of the gradient, and the discreteness is kept by the stochastic neuron. This approach has a similar presence to the reparametrization trick~\cite{vae}, and is unbiased and stable during training. This is also adopted in ~\cite{zsih}, and JMLH follows the same idea. An alternative simple choice here is the Straight-Through (ST) estimator~\cite{bengio2013estimating}, which is used in GreedyHash~\cite{greedyhash}. The REINFORCE algorithm~\cite{reinforce} is also employed for the same purpose in \cite{pgdh}, while it undergoes high variance during training.

\subsection{Enriching the Semantic Information}
JMLH is not the first model that trains the hashing network with classification objectives. For instance, SUBIC~\cite{subic} also employs a classification loss as its learning objective. Specifically, SUBIC~\cite{subic} separates the hash code into $l$ blocks and ground each code block on a $\Delta^{\frac{m}{l}-1}$ simplex in order to favor the discreteness. This approach considerably limits the maximal information carried by the codes. Besides, the supervised version of GreedyHash~\cite{greedyhash} is similar to JMLH both in terms of classification objective and keeping the discrete constraints. However, GreedyHash~\cite{greedyhash} only uses the quantization loss on the code bottleneck, ignoring the entropy of the codes, while we consider minimizing $\operatorname{KL}\left(q(\mathbf{b}|\mathbf{x})||\mathcal{B}(\mathbf{b}|m,0.5)\right)$ to preserve the entropy. Moreover, GreedyHash~\cite{greedyhash} provides no theoretical clue of how the trained codes are related to data semantics.

MIHash~\cite{mihash} borrows the concept of mutual information as with JMLH, ending up with different designs. Our model reflects the mutual information between codes and data semantics as a part of VIB~\cite{vib}, while MIHash~\cite{mihash} considers relevant-irrelevant code distribution discrepancy and requires complex histogram binning~\cite{NIPS2016_6464} during training.

Recently, a popular idea in deep representation learning is to employ Generative Adversarial Networks (GANs)~\cite{gan} during training, which has been attempted in \cite{shashgan,hashgan,bgan,bingan}. The discriminators or the encoders in GANs are aware of the data distribution $p(X)$ without explicitly parameterizing $p(X)$. The problem is that the auxiliary generator significantly increases the training complexity as more parameters are introduced. 

We experimentally show that the above sophisticated designs are not always necessarily needed as the simple network of JMLH can already achieve the state-of-the-art retrieval performance.

\section{Experiments}~\label{sec_4}
\begin{table*}[t]
	\begin{center}
		\caption{Performance comparison (\wrt mAP@$k$) of JMLH and the state-of-the-art hashing methods. The respective retrieval sequence length $k$ is adopted according to the most popular settings~\cite{hashgan,greedyhash,pgdh}. All baselines are reported according to the identical setting.%Note that TBH is \textcolor{blue!80!black}{\textbf{unsupervised}}. The retrieval performance of some supervised hashing methods is reported only for better illustration.
		}\vspace{0ex}
		\label{tab_map}
		\small
		\resizebox{.95\textwidth}{!}{
			\begin{tabular}{lc ccc ccc ccc}
				\hline
				%\rowcolor{gray!20}
				\multirow{2}{*}{\textbf{Method}}&\textbf{Super-}&\multicolumn{3}{c}{\textbf{CIFAR-10} (mAP@all)}&\multicolumn{3}{c}{\textbf{NUS-WIDE} (mAP@5000)}&\multicolumn{3}{c}{\textbf{ImageNet} (mAP@1000)}\\\cline{3-11}
				&\textbf{vision}& 16 bits& 32 bits & 64 bits& 16 bits& 32 bits & 64 bits& 16 bits& 32 bits & 64 bits\\ \hline\hline
				%KMH~\cite{he2013k}&\xmark& 0.279 & 0.296& 0.334 & 0.562& 0.597 & 0.600 & 0.543 & 0.554 & 0.592 & 0.202 & 0.297 & 0.390 \\
				%SpherH~\cite{sph}&\xmark& 0.254 & 0.291 & 0.333 & 0.495 & 0.558 & 0.582 & 0.516& 0.547 &0.589 & 0.110 & 0.187& 0.259\\
				% PCAH&\xmark& 21.52& 21.62& 20.54& & & & & & & & & \\
				%LSH~\cite{lsh}&\xmark& 0.106& 0.102 & 0.105 & 0.239& 0.266 & 0.266& 0.353& 0.372& 0.341& 0.152 & 0.141 & 0.165\\
				ITQ~\cite{itq}&\xmark& 0.201& 0.207& 0.235& 0.627& 0.645& 0.664& 0.217& 0.317& 0.391\\
				%SpH~\cite{sh}&\xmark& 0.272& 0.285& 0.300 & 0.517& 0.511& 0.510 & 0.527& 0.529& 0.546& 0.185& 0.271& 0.350\\
				AGH~\cite{agh}&\xmark& 0.217& 0.205 & 0.182 & 0.592& 0.615& 0.616 & 0.241& 0.327& 0.379\\
				DGH~\cite{dgh}&\xmark& 0.199& 0.200& 0.212& 0.572& 0.607& 0.627 &0.270 &0.341 &0.373\\
				% DH&\xmark& 16.17& 16.62& 16.96& & 48& & & & & & &\\
				%HashGAN~\cite{shashgan}&\xmark& 0.447& 0.463& 0.481& -& -& -& -& -& -& -& -& -\\
				%DeepBit~\cite{deepbit}&\xmark& 0.194& 0.249& 0.277& 0.392& 0.403& 0.429& 0.407& 0.419 & 0.430 & 0.204& 0.281 &0.286\\
				% BGAN~\cite{bgan}&\xmark& & 53.10& & & 71.4& & & & & & &\\
				%SGH~\cite{sgh}&\xmark & 0.435& 0.437 & 0.433& 0.593& 0.590& 0.607& 0.594& 0.610& 0.618& 0.447& 0.500& 0.523\\
				%GreedyHash~\cite{greedyhash}&\xmark& 0.448& 0.473& 0.501& 0.633& 0.691& 0.731 & 0.582& 0.668& 0.710& 0.186& 0.578& 0.558\\\hline
				%\rowcolor{blue!15}
				%\textbf{TBH (Ours)}&\xmark&\textbf{0.532}&\textbf{0.573}&\textbf{0.578}&\textbf{0.717}&\textbf{0.725}&\textbf{0.735}&\textbf{0.706}&\textbf{0.735}&\textbf{0.722}&\textbf{0.560}&\textbf{0.619}&\textbf{0.626}\\\hline\hline
				KSH~\cite{ksh}&\checkmark& 0.451 & 0.473 & 0.507 & 0.448 & 0.520& 0.566 & 0.216 & 0.257 & 0.394\\
				ITQ-CCA~\cite{ccaitq}&\checkmark& 0.463& 0.498& 0.505& 0.555& 0.512& 0.460 & 0.235& 0.377& 0.576\\
				SDH~\cite{sdh}&\checkmark& 0.499& 0.525& 0.546 & 0.595& 0.595& 0.617& 0.298& 0.431& 0.504\\\cline{1-11}
				CNNH~\cite{cnnh}&\checkmark& 0.453 & 0.509& 0.537 & 0.570& 0.583& 0.600& 0.281& 0.450& 0.554\\
				DNNH~\cite{dnnh}&\checkmark& 0.556& 0.558& 0.599& 0.598& 0.616& 0.639& 0.290& 0.461& 0.565\\
				DHN~\cite{dhn}&\checkmark& 0.564& 0.603& 0.626& 0.637& 0.664& 0.671& 0.311& 0.472& 0.573\\
				HashNet~\cite{cao2017hashnet}&\checkmark&0.643 &0.675 &0.687 & 0.662& 0.699& 0.716& 0.506& 0.631& 0.684\\
				MIHash~\cite{mihash}&$\checkmark$& 0.760& 0.776& 0.761& 0.722& 0.759& 0.779&0.569&0.661&0.694\\
				HashGAN~\cite{shashgan}&\checkmark& 0.668& 0.731& 0.749& 0.715& 0.737& 0.748& -& -& -\\
				PGDH~\cite{pgdh}&\checkmark &0.741 &0.747 &0.762 &0.780 &0.786 &0.792 &0.653 & 0.707 & 0.716\\
				GreedyHash~\cite{greedyhash}&\checkmark& 0.786& 0.810& 0.833& -& -&- & 0.625& 0.662& 0.688\\\hline
				\textbf{JMLH (Ours)} &\checkmark& \textbf{0.805}& \textbf{0.841}& \textbf{0.837}& \textbf{0.795}& \textbf{0.818}& \textbf{0.820}& \textbf{0.668}& \textbf{0.714}& \textbf{0.727}\\\hline
			\end{tabular}
		}
	\end{center}
\end{table*}%\vspace{-2ex}

Extensive image retrieval experiments are conducted in this section, mainly according to the following themes:
\begin{itemize}
	\item \textbf{Comparison with existing methods.} We show that, simple as JMLH is, it still outperforms state-of-the-art hashing models.
	\item \textbf{Ablation study.} The importance of each part of JMLH is evaluated and discussed.
	\item \textbf{Intuitive results.} Some illustrative results are provided to implicitly justify the effectiveness of JMLH.
\end{itemize}
\subsection{Experimental Settings}
\subsubsection{Implementation Details}

JMLH is implemented with the popular deep learning toolbox Tensorflow~\cite{tf}. The network specifics are provided in Table~\ref{tab_net}. 
For our image retrieval task, AlexNet~\cite{alexnet} before the \texttt{fc\_7} layer is adopted as the network backbone, where parameters are initialized with the ImageNet~\cite{imagenet} pre-trained results and is jointly updated during training. For multi-labeled datasets, \ie, NUS-WIDE~\cite{nus}, we activates the last layer of $\pi(y|\mathbf{b})$ with the \texttt{sigmoid} non-linearity, while the \texttt{softmax} activation is used when training JMLH on CIFAR-10~\cite{cifar} and ImageNet~\cite{imagenet}. 
JMLH involves one hyper-parameter, \ie, the regularization factor $\lambda$. We empirically set $\lambda=0.1$. The learning rate of the Adam optimizer $\mathbf{\Gamma}\left(\cdot\right)$~\cite{adam} is set to $1\times10^{-4}$. We fix the training batch size to 256. The codes can be found at \url{https://github.com/ymcidence/JMLH}.
\subsubsection{Datasets}
\noindent\textbf{CIFAR-10~\cite{cifar}} consists of 60,000 images from 10 classes. We follow the common setting \cite{hashgan,dnnh,greedyhash} and select 1,000 images (100 per class) as the query set. The remaining 59,000 images are regarded as the database. The training set contains 5000 images, uniformly selected from the database.

\noindent\textbf{NUS-WIDE~\cite{nus}} is a collection of nearly 270,000 Web images of 81 categories downloaded from Flickr. Following the settings in \cite{agh,cnnh,dnnh}, we adopt the subset of images from the 21 most frequent categories. 100 images of each class are utilized as a query set and the remaining images form the database. For training, we employ 10,500 images uniformly selected from the 21 classes.

\noindent\textbf{ImageNet~\cite{imagenet}} is originally released for large-scale image classification purpose, and is recently used in deep hashing evaluation. 
%in Large Scale Visual Recognition Challenge (ILSVRC). 
Following \cite{cao2017hashnet,pgdh}, we randomly select 100 categories to perform our retrieval task. All the original training images are used as the database, and all the validation images form the query set. For each category, 130 images are used for training.

\subsection{Comparison with Existing Methods}
We compare JMLH with existing methods using conventional evaluation metrics, including top-$k$ mean-Average Precision (mAP@$k$), Precision of top-$k$ retrieved samples (Precision@$k$), Precision within Hamming radius of 2 (P@H$\leq$2) and Precision-Recall (P-R) curves.

Note that, for mAP@$k$, we adopt the most popular settings of $k=all, 5000, 1000$ for \textbf{CIFAR-10}, \textbf{NUS-WIDE}, and \textbf{ImageNet} respectively according to \cite{hashgan,greedyhash,pgdh}.
\subsubsection{Baselines}
JMLH is compared with various widely recognized hashing baselines, including ITQ~\cite{itq}, AGH~\cite{agh}, DGH~\cite{dgh}, KSH~\cite{ksh}, ITQ-CCA~\cite{ccaitq}, SDH~\cite{sdh}, CNNH~\cite{cnnh}, DNNH~\cite{dnnh}, DHN~\cite{dhn}, HashNet~\cite{cao2017hashnet}, HashGAN~\cite{shashgan} PGDH~\cite{pgdh} and the supervised version of GreedyHash~\cite{greedyhash}. Note that the term of \textit{HashGAN} is used both in \cite{hashgan} and \cite{shashgan}. Here we refer to the later one as it is a supervised approach and thus is more related to our work.

For feature-based models, \eg, shallow hashing models, we use the AlexNet~\cite{alexnet} \texttt{fc\_7} pre-trained features to represent data for training and test. As for the end-to-end baseline frameworks, we directly adopt the original training settings described in their original papers and pre-trained weights are also applied for fine-tuning when possible.
\subsubsection{Results and Analysis}
\begin{figure*}[t]
	\begin{center}
		\includegraphics[width=\textwidth]{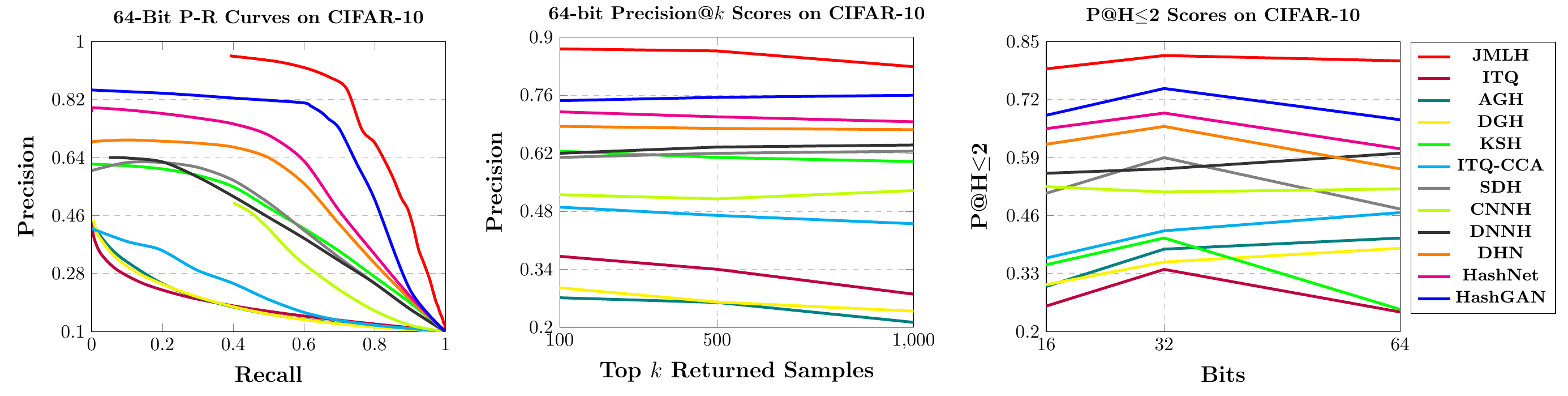}
	\end{center}
	\caption{\textbf{Left:} 64-bit P-R curves on CIFAR-10~\cite{cifar}. \textbf{Middle:} 64-bit precision of top $k$ returned samples on CIFAR-10~\cite{cifar}. \textbf{Right:} Precision within Hamming radius of 2 scores on CIFAR-10~\cite{cifar}.}
	\label{fig_pr}
\end{figure*}

The retrieval mAP@$k$ results are reported in Table~\ref{tab_map}. The respective P-R curves, Precision@$k$ and P@H$\leq$2 scores are illustrated in Figure~\ref{fig_pr}.

It can be observed that JMLH consistently outperforms the compared baselines, though many of them consist of more trainable parameters, \eg, HashGAN~\cite{shashgan}. This result aligns with our motivation, and shows the clue that, with the current evaluation metrics, one may not require an extremely complex model to obtain the best-performing deep hashing function.

The performance margin between JMLH and GreedyHash~\cite{greedyhash} is not significant on CIFAR-10~\cite{cifar}, but this gap gets larger when it comes to a relatively more difficult situation, \ie, ImageNet~\cite{imagenet}. This raises the concern of a proper regularization term for training. Both GreedyHash~\cite{greedyhash} and JMLH are trained with classification-oriented objectives. The former literally involves a quantization penalty while JMLH considers equally distributed $\{0, 1\}$ bits to maximize the expected code entropy. This factor becomes essential when the data label space is large and the training samples are limited as the codes need to be expressive enough to be successfully classified. We find our design has better generalization ability in this case.

\subsection{Ablation Study}

In this subsection, we evaluate different components in terms of formulating a simple deep hashing model, and empirically show which one is of importance for good performance.
\subsubsection{Baselines}
%The ablation baselines of JMLH are introduced as follows. Note that some of them could be similar to existing models.

\noindent \textbf{JMLH-Cont.} We firstly look at the influence of quantization. By dropping the binary stochastic neuron and employing the \texttt{sigmoid} activation on the code bottleneck $B$, a regular deep neural classifier is built. The regularization term is kept, and is subsequently analyzed by other baselines.

\noindent \textbf{JMLH-QR.} The $\operatorname{KL}$ term of Eq.~(\ref{eq_3}) is replaced by the \textbf{quantization regularizer} between the activated binary codes $B$ and their real-valued counterparts before the stochastic neurons.

\noindent \textbf{JMLH-NR.} The regularizer is deprecated in this baseline, and the whole learning objective is formulated by the classification cross-entropy.

\noindent \textbf{JMLH-VAE.} We replace the classifier $\pi(\cdot)$ with a decoder, and use the $L2$ reconstruction error instead of classification loss during training. Therefore, the model collapses to an unsupervised Variational Auto-Encoder (VAE)~\cite{vae}, with a negative Evidence Lower-BOund (ELBO) of
\begin{equation}
\frac{1}{n}\sum_{\mathbf{x}}\mathbb{E}_{q(\mathbf{b}|\mathbf{x})}[-\log q(\mathbf{x}|\mathbf{b})]
+ \operatorname{KL}\left(q(\mathbf{b}|\mathbf{x})||p(\mathbf{b})\right).
\end{equation}
For the simplicity of training, the encoder and decoder for this baseline are both implemented with a two-layer neural networks and are fed by AlexNet~\cite{alexnet} \texttt{fc\_7} features.

\subsubsection{Results and Analysis}
\begin{table}
	\caption{mAP@all results by using different variants of the proposed JMLH on CIFAR-10.}
	\label{tab_abl}
	\small
	\resizebox{0.99\linewidth}{!}{
		\begin{tabular}{ll ccc}
			\hline
			&\textbf{Baseline}& \textbf{16 bits}& \textbf{32 bits}& \textbf{64 bits}\\\hline\hline
			1&JMLH-Cont & 0.616& 0.628 & 0.659\\
			2&JMLH-QR & 0.778 & 0.827 & 0.835\\
			3&JMLH-NR & 0.729& 0.725 & 0.736\\
			4&JMLH-VAE & 0.423& 0.435& 0.441\\\hline
			5&\textbf{JMLH (full model)}& \textbf{0.805}& \textbf{0.841}& \textbf{0.837}\\\hline
		\end{tabular}
	}
	\vspace{-2ex}
\end{table}

The mAP results of the above-mentioned baselines are shown in Table~\ref{tab_abl}. Since JMLH-VAE is an unsupervised model, its performance is relatively lower than the others. We experience a 20\% performance drop when using the continuous relaxation during training, \ie, JMLH-Cont. As discussed in Section~\ref{sec_3}, the binary constraints are essential for models like JMLH as it directly influences the classifier's observation. Without regularization, JMLH-NR struggles in the training-test generalization. Though not competing our full model, JMLH-QR still performs closely to GreedyHash~\cite{greedyhash}, as the learning objectives are similar. The difference between JMLH-QR and GreedyHash~\cite{greedyhash} lies in the stochasticity of gradient estimation. Both ST~\cite{bengio2013estimating} and distributional derivative~\cite{sgh} work for this case as long as the binary constraints are not violated. Hence, a proper learning objective becomes more important.

\subsection{More Results}

\subsubsection{Hyper-Parameter}
\begin{figure}[t]
	\begin{center}
		\includegraphics[width=\linewidth]{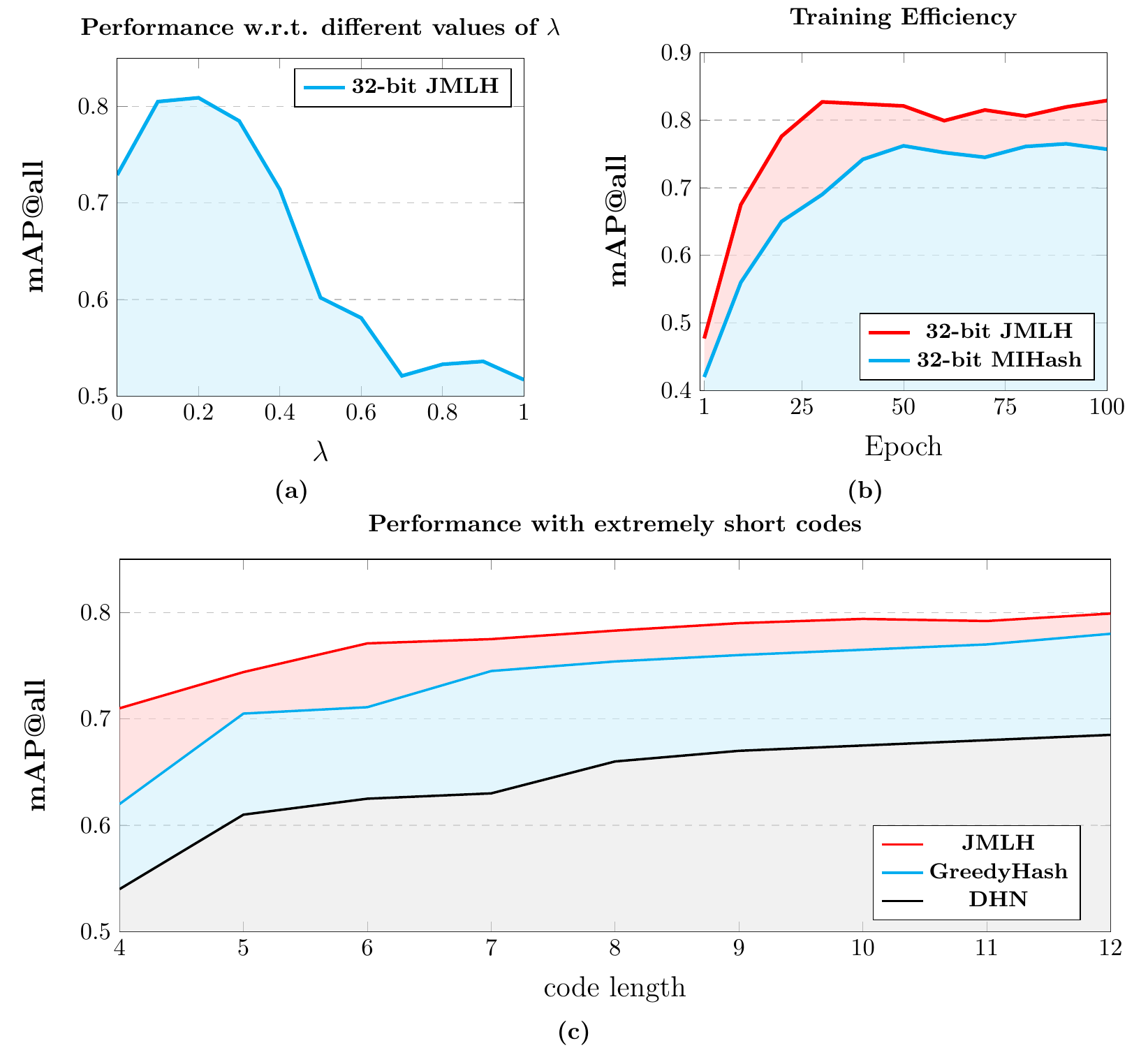}
	\end{center}
	\caption{\textbf{(a)} mAP@all results of 32-bit JMLH on CIFAR-10~\cite{cifar} with different values of $\lambda$. \textbf{(b)} Training efficiency of JMLH and MIHash~\cite{mihash} on CIFAR-10~\cite{cifar}. \textbf{(c)} Encoding performance comparison with extremely short code length on CIFAR-10~\cite{cifar}.}
	\label{fig_hyper}
\end{figure}

The regularization penalty of JMLH is scaled by a hyper-parameter $\lambda$. By default, it is set to $\lambda=0.1$ for the overall best performance. The impact of $\lambda$ is illustrated in Figure~\ref{fig_hyper} (a). The performance drops quickly when $\lambda$ goes larger, which actually reflects the penalty of the mutual information between data $X$ and codes $B$, \ie, $I(X,B)$. A large value of $\lambda$ over-regularizes the model by decorrelating $X$ with $B$, making the produced codes less-informative.

\subsubsection{Towards Model Simplicity}
One key claim of this paper is to build a simple deep hashing model. Training JMLH is non-trivial and efficient. Our classification likelihood learning objective provides a straightforward way to convey data semantics to the encoder. We show training efficiency comparison between JMLH and MIHash~\cite{mihash} in Figure~\ref{fig_hyper} (b). It can be observed that JMLH converges more quickly to the best performance than MIHash~\cite{mihash} with a margin of $\sim$10 epochs. Although MIHash \cite{mihash} requires no auxiliary networks, its histogram-based learning objective introduces complex positive-negative data pairing and histogram binning. All these factors make the training of MIHash~\cite{mihash} indirect, resulting in relatively slower convergence rate than JMLH. Note that the performance of MIHash is slightly lower than the one reported in~\cite{mihash}, as it was previously trained with VGG~\cite{vgg} features and we reproduce the results with the AlexNet~\cite{alexnet} backbone for fair comparison.

The whole parameter size of JMLH for all experiments conducted in this section is slightly smaller than AlexNet~\cite{alexnet}, as we have a relatively narrow fully-connecting bottleneck in the middle. Compared with the models that involve end-to-end generative networks~\cite{hashgan,shashgan}, this is believed to be a light one.

\subsubsection{Extremely Short Codes}
Following \cite{greedyhash}, we also explore the minimal size of codes to represent data semantics. The experiments are conducted by setting the code length to $m=4,5,...,11,12$, and the corresponding results are shown in Figure~\ref{fig_hyper} (c). We can see that, compared with GreedyHash~\cite{greedyhash} and DHN~\cite{dhn}, JMLH obtains better performance even when the encoding length is very short. The entropy-preserving regularization term plays the key role here since the maximum number of concepts that the code space can cover is limited.
\subsubsection{Visualization Results}
\begin{figure}[t]
	\begin{center}
		\includegraphics[width=\linewidth]{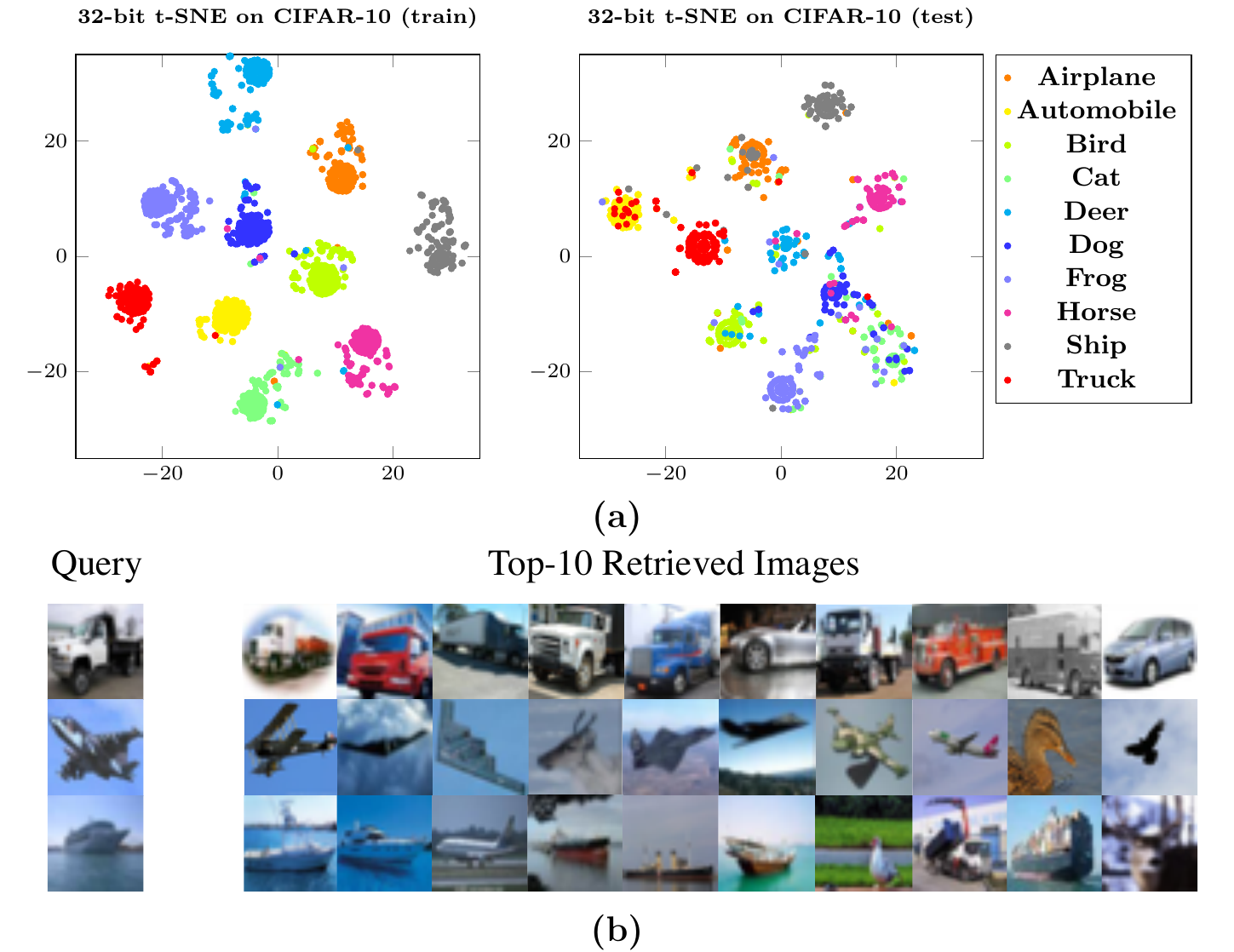}
	\end{center}
	\caption{\textbf{(a)} 32-bit JMLH t-SNE~\cite{tsne} visualization on CIFAR-10~\cite{cifar}. \textbf{(b)} Examples of top-10 retrieved candidates of 32-bit JMLH on CIFAR-10~\cite{cifar}.}
	\label{fig_tsne}
\end{figure}

The t-SNE~\cite{tsne} visualization of 32-bit JMLH on CIFAR-10~\cite{cifar} is shown in Figure~\ref{fig_tsne} (a). Even though the proposed model is simple both in terms of network structure and learning objective, the resulting binary codes are still clearly scattered in the feature space according to their semantic meanings. We further provide several image retrieval examples where the top-10 retrieved candidates are shown together with the query image in Figure~\ref{fig_tsne} (b). Obviously, JMLH successfully finds related images in the top of the retrieval list. Here we only show the 32-bit results to keep the content concise.

%\subsection{Discussion}
%Upon the above experiments, we motivate this paper by discussing the essential issues for designing a well-performing but simple binary representation learning model, and thus further elaborate the possibility of improvement and future work.
%
%We empirically summarize the key factors for a good hashing model as follows:
%\begin{itemize}
%	\item \textbf{The binary constraints during training are important.} This typically applies to the occasion where a successive semantic-learning network is topped on the binary encoding layer. One can observe significant performance difference between JMLH and JMLH-Cont.
%	\item \textbf{text}
%\end{itemize}

\section{Conclusion}~\label{sec_5}
In this paper, we proposed a simple but effective deep hashing model called JMLH. Our model shaped a conventional deep neural network with a single likelihood maximization learning objective. A differentiable binary bottleneck was plugged in, making the whole network end-to-end trainable using SGD. JMLH was linked to the information bottleneck methods, which aimed at learning maximally representative features for a given task. We showed that, when applying proper binary-preserving gradient estimators and suitable regularization terms, a single classification model could generate high-quality hash codes for similarity search, outperforming state-of-the-art models.

{\small
	\bibliographystyle{ieee}
	\bibliography{refs}
}

\end{document}